\documentclass[10pt]{article}

\usepackage{amsmath, amsfonts, amssymb, mathrsfs, amsthm}
\usepackage{lineno}
\usepackage{hyperref}

\usepackage{braket}

\modulolinenumbers[5]

\usepackage{wrapfig}

\def\drawplusplus#1#2#3{\hbox to 0pt{\hbox to #1{\hfill\vrule height #3 depth
      0pt width #2\hfill\vrule height #3 depth 0pt width #2\hfill
      }}\vbox to #3{\vfill\hrule height #2 depth 0pt width
      #1 \vfill}}

\usepackage[all]{xy}
\usepackage{graphicx}

\newtheorem{mydef}{Definition}

\newtheorem*{thm*}{Theorem}
\newtheorem*{mydef*}{Definition}
\newtheorem*{mylemma*}{Lemma}
\newtheorem*{myconjecture*}{Conjecture}

\begin{document}


\title{Paraconsistent Foundations\\
for Probabilistic Reasoning, Programming \\
and Concept Formation}

\author{Ben Goertzel}





\maketitle

\begin{abstract}
It is argued that 4-valued paraconsistent truth values (called here {\it p-bits}) can serve as a conceptual, mathematical and practical foundation for highly AI-relevant forms of probabilistic logic and probabilistic programming and concept formation.   

First it is shown that appropriate averaging-across-situations and renormalization of 4-valued p-bits operating in accordance with Constructible Duality (CD) logic yields PLN (Probabilistic Logic Networks) strength-and-confidence truth values.  Then variations on the Curry-Howard correspondence are used to map these paraconsistent and probabilistic logics into probabilistic types suitable for use within dependent type based programming languages.   

Zach Weber's paraconsistent analysis of the sorites paradox is extended to form a paraconsistent / probabilistic / fuzzy analysis of concept boundaries; and a paraconsistent version of concept formation via Formal Concept Analysis is presented, building on a definition of fuzzy property-value degrees in terms of relative entropy on paraconsistent probability distributions.  

These general points are fleshed out via reference to the realization of probabilistic reasoning and programming and concept formation in the OpenCog AGI framework which is centered on collaborative multi-algorithm updating of a common knowledge metagraph.
\end{abstract}



\tableofcontents

\section{Introduction}

In standard logic systems, falsehood is explosive -- meaning that once a single falsehood creeps into a knowledge-base, then straightforward logical inference will lead that knowledge-base to include all possible statements expressible in its logical language as both truths and falsehoods.   So once you have accepted a single falsehood, you're totally bollixed and basically can't reason effectively about anything anymore.

Naively, this seems not to be how human minds reason.   People seem able to embrace {\it some} falsehoods and contradictions, without as a result embracing {\it all} falsehoods and contradictions.   

One could take this situation as evidence that logic is a poor model of human cognition and reasoning.

However, this would be an unwarranted generalization, because there are logic systems in which falsehoods are not explosive -- these are the paraconsistent logics.   In a paraconsistent logic, inconsistency is a tractable and manageable aspect of reasoning along with various others.

So far, paraconsistent logic has not played a huge role in logic-based AI or cognitive modeling.   However, we believe that it should, and toward that end we here use a version of paraconsistent logic to provide new formal underpinnings for concepts and methods in probabilistic reasoning and programming and concept formation that relate to our current work on integrative AGI according to the OpenCog paradigm \cite{EGI1} \cite{EGI2}..

We begin with Anna Patterson's Constructible Duality logic (which we here call CD logic), which has four truth values: True, False, Neither True nor False, Both True and False.   We label these 4-valued truth values {\it p-bits},  and show how CD p-bit logic, appropriately extended, helps provide a formal foundation for the PLN probabilistic logic framework used in the OpenCog AI system.   We then show how the Curry-Howard-correspondent image of CD logic manifests itself in the realm of dependent type theory.

Zach Weber \cite{weber2010paraconsistent} has given an elegant analysis of the sorites paradox (Since if $n$ grains of wheat is not enough to make a heap, then $n+1$ is not either, it follows that one billion grains of wheat is not enough to make a heap), which we extend here into a theory of concept boundaries that pulls together paraconsistent, probabilistic and fuzzy logic, along with dependent type theory.    Finally, we show how Formal Concept Analysis (FCA) can be extended to derive concept lattices in the context of properties whose relationships to objects are quantified with paraconsistent truth values.   

As we proceed through these abstract issues, we will here and there flesh out the concepts presented via examples drawn from our work with the OpenCog system \cite{EGI1}\cite{EGI2}, in which PLN, probabilistic programming, concept formation and other associated mechanisms are defined with respect to a comprehensive dynamic knowledge metagraph; in these aspects some recent work on the properties of metagraphs \cite{Goertzel2020metagraph} is relevant.

We believe that logic systems have great promise for formalizing key aspects of human reasoning as well as for structuring significant portions of the reasoning of narrow AI and AGI systems; however, in order for this promise to be realized, an appropriately flexible vision of the scope of logic systems must be adopted, and mappings between declarative knowledge and other sorts of knowledge must be attended.

\section{Connecting Probabilistic Logic with Paraconsistent Logic}
\label{sec:CD-logic}

If one wishes to configure  a logic that can incorporate and manage inconsistent pieces of evidence and the inconsistent conclusions that naturally follow from them, there are multiple routes to pursue, and these interconnect in some cases subtly.

The Probabilistic Logic Networks \cite{PLN} framework leveraged in OpenCog allows inconsistent pieces and bodies of evidence, and contains tools for managing this inconsistency, grounded in a rigorously observation-based semantics.   Paraconsistent logic manages inconsistency in a different way, often leveraging extensions of constructive/intuitionistic logic, which has an observation based semantics of a related kind.   Here we build a precise mapping between PLN and a certain type of paraconsistent logic, Constructive Duality (CD) logic.   

A Constructive Duality logic consists of two "standard" constructive logics paired in a certain way, and there is a well known Curry-Howard type mapping from constructive logic into dependent type based programming languages.   The relation between CD logic and PLN thus gives us a basis for mapping PLN into dependent type based programming with probabilistic types --  bringing us onward to the topic of Sections \ref{sec:types-background} and \ref{sec:types} below.

\subsection{Probabilizing Logics}

A wide variety of crisp logics can be mapped directly into corresponding probabilistic logics by adopting the right sort of possible-worlds semantics (actually we will frame it as "possible-situations semantics", but there is strong overlap with notions of possible-worlds semantics in the literature as well as with situation semantics \cite{Barwise1989a}).   This is particularly straightforward in the case of intuitionistic logics, which are the variety that most interest us here, but the same process could be extended more broadly, with limits whose precise delineation are yet to be explored.   

We will consider here the broadly defined case of logic that assigns truth values, drawn from some range (which could be finite, countable or infinite depending on the case), to various propositions regarding logical atoms, where the logical atoms are viewed as observations of some world.   Given this very broad setting we will explain how to create a probabilistic logic extending the logic in question.

To have a simple model of "observations" and "situations", we can assume 

\begin{itemize}
\item there is some countable set of properties, each of which has some countable set of possible values
\item each observation is a set of (property, value) pairs
\item a situation is a set of "possible observations"
\end{itemize} 

\noindent One can modify this basic framework in various ways without affecting the overall conclusions here significantly (e.g. introducing fuzzy sets of various sorts, looking at well-behaved uncountable sets, restricting things to finite sets, etc.)

In this framework, the atomic logical entities ("Atoms") $\mathcal{A}$ are then:

\begin{itemize}
\item Observation $O_i$ was made.
\item Observation $O_i$ has value $v_{ij}$ for feature $F_j$
\end{itemize}
 
 \noindent We will be concerned here with logics  $\mathcal{L}$ that include
 
 \begin{itemize}
\item a formal language involving  
 \begin{itemize}
 \item  tokens directly referring to entities in $\mathcal{A}$ 
\item variables whose values are entities in  $\mathcal{A}$ 
\item symbols serving as connectors between these tokens and variables
\end{itemize}
\item a function that assigns, to expressions in this language (which we will call "propositions"), truth values drawn from a particular set
\end{itemize}

\noindent It's OK if some propositions expressible in the language aren't assigned truth values by the truth-value function; we can consider this as assignment of a null truth value.

Now let us suppose we have a large set $\mathcal{S}$ of possible situations, all defined over the same set of possible properties and values.   In each situation, we may assume, a different set of observations is made.   

Given a probability distribution $\Phi$ over possible situations in $\mathcal{S}$ , and a logical proposition $S$ defined using logic $\mathcal{L}$ over Atoms in $\mathcal{A}$, one can then calculate

$$
t_\Phi (S)
$$  

\noindent , the expected truth value of $S$ for a possible situation randomly selected from $\mathcal{S}$.\footnote{This assumes of course that truth values are drawn from a space where expected values can be calculated.}    One can also define a whole probability distribution $\rho_\Phi(S)$, of which $t_\Phi (S)$ is the mean.

\subsubsection{Creating the Situation Ensemble}

One may question where, in a AGI or cognitive science context, all these possible situations come from.   In a sense, an intelligent agent is going on its own experience, its own current and remembered perceptions -- which are most simplistically conceived as representing "one world" containing one set of remembered situations, which are in a sense definite and not merely possible.   However, things get less simplistic when one incorporates the fundamental uncertainty associated with any perception, observation or memory.   Incorporating this aspect, one can articulate and formally model an ensemble of possible situations corresponding to a single intelligent agent's experience, including situations that are not actually realized as part of the agent's history according to its episodic memory, nor believed to have likely ever existed according to its inferences.

For instance, one can create such possible situations via:

\begin{itemize}
\item  removing a randomly selected collection of experiences from the agent's memory
\item spatially or temporally reorganizing portions of the agent's experience
\item  more generally and encompassing the prior two options: assuming a counterfactual
hypothesis (i.e. assigning a statement a truth value
that contradicts the agents experience), and using
inference to construct a set of observations that is
as consilient with the agent's memory as possible, subject to the
constraint of being consistent with the hypothesis.
\end{itemize}

\noindent In \cite{ikle2010grounding} the first option, which connects closely with the theory of subsampling in statistics, is explored in some mathematical details.

Putting these pieces together, we then have a way of probabilizing a logic relative to  particular agent with perception and memory.   We have treated the logic $\mathcal{L}$ itself in an extremely simplistic way, however in the case of constructive/intuitionistic logics (our main interest) this seems unproblematic.

The probabilistic logic $\mathcal{P}_\mathcal{L}$ constructed here has an obvious relationship to the base crisp logic  $\mathcal{L}$; i.e. it reduces precisely to $\mathcal{L}$ for the case of propositions that have the same truth value independently of what situation in $\mathcal{S}$  is chosen, and it reduces to $\mathcal{L}$ for all propositions in the case $\Phi$ is a delta function focused on just one situation .

\subsection{Constructible Duality Logic}

The variant of formal logic that appears to best match the natural semantics of experiential learning based AI systems is the constructible duality logic articulated by Patterson \cite{patterson1998implicit}.

Constructible duality (CD) logic begins by following earlier work on constructible falsity \cite{nelson1949constructible}, which improves on the notion of negation used in standard intuitionistic logic.   In CD as in constructible falsity based logics, negation of $S$ does not mean showing that assuming $S$  yields an absurdity, but rather means coming up with a specific counterexample construction showing $S$  is false.   This remedies an asymmetry between truth and falsehood that exists in standard intuitionistic logic.

\subsubsection{Species of Negation}

Classical logic differs from standard intuitionistic logic via including the excluded-middle rule $A \lor \neg A$.  Constructible falsity logic differs from standard intuitionistic logic via requiring an effective method for constructing counterexamples to $A$, as opposed to just requiring a proof that assuming $A$ leads to a contradiction via application of inference rules.   

For instance, if $A= \forall x f(x)$ then a constructive refutation would be a specific demonstration of some $n$ for which $f(n)$ is constructively proven true.   If  $A= \exists x f(x)$ then a constructive refutation would consist of specific demonstrations for some countable set $n_i$ that $\neg f(n_i)$, combined with a procedure $F$, that for any $m$, can be constructively proved to take any proof that $f(m)$ and produce a proof that $f(n_i)$ for some $i$.  \footnote{The $ \exists $ case is not discussed as explicitly or extensively in the literature, so the discussion of this case here is somewhat improvised and in need of further formal elaboration.   }

There is close connection between constructible falsity and Popper's philosophy of science with its focus on falsification.  Shramko \cite{shramko2005} has proposed a logic of falsifications, which is dual to standard intuitionistic logic in the sense that it requires constructions to demonstrate falsehood (i.e. specific evidentiary examples to show an hypothesis false) and then allows truth to be shown indirectly, via proving that assuming $\neg A$ leads to a contradiction via application of inference rules.

In these three systems, then,

\begin{itemize}
\item  In falsification logic, "false" means "I have counterexamples" whereas "true"  means "can't assume it's false."
\item In standard intuitionistic logic, "true" means "I have examples" whereas "false" means "can't assume it's true."
\item In constructible falsity logic "true" means "I have examples" and false means "I have counterexamples."
\end{itemize}

\noindent In \cite{shramko2005} these three forms of negation are placed in a more general context (an elegant "unified kite of negations") along with a variety of other negation operators.

\subsubsection{Constructible Duality Logic Operators}

Constructible falsity, as shown following the methods from Rasiowa \cite{rasiowa1974algebraic},  leads naturally to a substitutable implication defined as 

$$
A \Rightarrow B = (A \rightarrow B) \land ( \neg B \rightarrow \neg A) 
$$

\noindent Patterson then introduces an operator which acts together with this implication as a natural conjunction operator ,

$$
A \bigotimes B = \neg (A \Rightarrow \neg B)
$$

\noindent which obeys the deduction rule

$$
(A \bigotimes B ) \Rightarrow C = A \Rightarrow (B \Rightarrow C)
$$

\noindent and has the DeMorgan dual

$$
A \bigoplus B = \neg (\neg A \bigotimes \neg B) = \neg A \Rightarrow B
$$

Patterson observes that the natural identity for the $\bigotimes$ operator is neither True nor False, but rather an additional truth value which she calls Overdefined or $I$.   Basically the $I$ truth value connotes "both true and false", and only works without trivializing the logic if one removes the consistency axiom $A \land \neg A \rightarrow B$.    She also adds a constant $-$ connoting "neither true nor false", which like $I$ is equal to its own negation.

CD logic thus provides one route to what we will call {\it p-bits} -- "paraconsistent bits", each of which has 4 possible values: True, False, Both or Neither.  

Given any logical language featuring constructs for True, False, $\land$, $\lor$, $\rightarrow$ and $\neg$, and a mapping from expressions in the language into a Heyting algebra $\mathcal{H}$, the Heyting algebra operations of meet, join and complement form an intuitionistic logic.   Patterson shows that, similarly, if one has a mapping $h$ from expressions in the language into the product algebra $\mathcal{H} \times \mathcal{H}^{op}$ (where $\mathcal{H}^{op}$ denotes the opposite algebra to $\mathcal{H}$), then one obtains a CD logic with rules as follows.

\begin{itemize}
\item Basic mapping of logic operations
\begin{itemize}
\item $h(\alpha \land \beta) = h(\alpha) \sqcap h(\beta)$
\item $h(\alpha \lor \beta) = h(\alpha) \sqcup h(\beta)$
\item $h(\alpha \rightarrow \beta) = h(\alpha) \rightarrow h(\beta)$
\item $h(\alpha ) = \neg h(\alpha)$
\end{itemize}
\item Mapping of Four Units
\begin{itemize}
\item $h(1) = (1,0)$
\item $h(0) = (0,1)$
\item $h(-) = (0,0)$
\item $h(I) = (1,1)$
\end{itemize}
\item  Logical operation across coordinates
\begin{itemize}
\item $(x, x')  \sqcap (y,y') = (x \sqcap y, x' \sqcup y')$
\item $(x, x')  \sqcup (y,y') = (x \sqcup y, x' \sqcap y')$
\item $(x, x')  \rightarrow (y,y') = (x \rightarrow y, x' \sqcap y')$
\item $\neg (x, x')  =  (x',x) $
\end{itemize}
\end{itemize}

\noindent In a logic of this nature, the derivable logical formulas $\Psi$ are those that result in $h(\Psi) = (1,0)$ or $h(\Psi) = (1,1)$.

The "exponential" operator $!A = A \land I$ removes the negative content of $A$, making $A$'s falsity true and yielding elegant rules such as

\begin{itemize}
\item $A \rightarrow B = !A \Rightarrow B$
\item $!A \Rightarrow A$
\item $!!A = A$
\end{itemize}

\noindent Clearly ! maps  $\mathcal{H} \times \mathcal{H}^{op}$ into $\mathcal{H} \times  1$ which is isomorphic to  $\mathcal{H}$. 

The dual exponential operator $\Gamma A = A \lor I = \neg ! \neg A$ obeys similar rules

\begin{itemize}
\item $\neg B \rightarrow \neg A = A \Rightarrow \Gamma B $
\item $A \Rightarrow  \Gamma A$
\item $\Gamma A = \Gamma \Gamma A$
\end{itemize}

\noindent and the weak exponentials $!`A = A \land - $ and its DeMorgan dual $?` A = A \lor -$ have similar properties as well.

Patterson \cite{patterson2000exponentials} shows that this CD logic maps naturally into linear logic, a fascinating result which shows that the basic effect of the restriction on re-use of evidence that occurs in linear logic can be achieved indirectly via restricting negations to be constructible (though there are some interesting details, see Theorem 26 in \cite{patterson2000exponentials} \footnote{Basically, linear-logic formulas are generally valid in CD logic, and one gets equivalence between multiplicative linear logic validity and CD validity if one assumes the multiplicative linear logic formulas are valid in a certain sense}).
 
\begin{figure}
\centering
  \includegraphics[width=7cm]{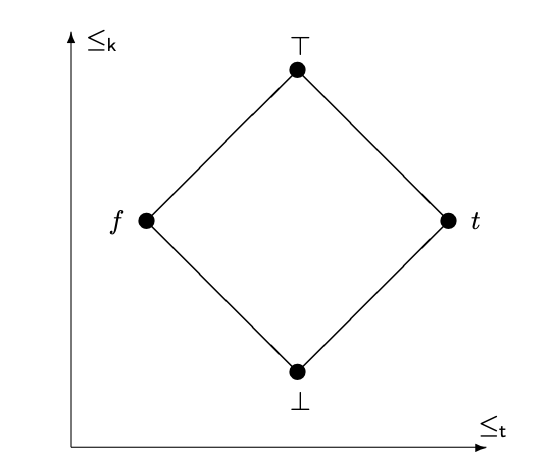}
  \caption{Belnap's depiction of the 4 truth values, with (1,0) at the right, (0,1) at the left, (0,0) at the bottom and (1,1) at the top.   The x-axis is an ordering based on amount of knowledge (confidence or count in PLN), and the y-axis is an ordering based on amount of truth (strength in PLN).}
  \label{fig:belnap}
\end{figure}

CD logic can be interpreted as a "Four Valued Paradefinite Logic" \cite{Belnap1977} \cite{arieli2017four}; Belknap analyzed the relationship between the four true values as falling into two lattices (see Figure \ref{fig:belnap}), one of which ranks them via amount of truth involved, the other by amount of evidence involved.

\subsubsection{CD Logic and Laws of Form}

Kauffmann and Colllings \cite{kauffman2019bf} show that the logical relationships of the core 4-valued (p-bit) truth values underlying CD logic can be generated from just two operations: conjunction and demi-negation $\upharpoonleft$, where the latter is a "square root of not" operator satisfying $A = \neg A$, defined on CD-style two-component truth values as

$$
\upharpoonleft (x,x') = (\neg x', x)
$$
                               
\noindent so that

\begin{itemize}
\item $\upharpoonleft (0,0) = (1,0) $
\item $\upharpoonleft (1,0) = (1,1)$ 
\item $\upharpoonleft (1,1) = (0,1) $
\item $\upharpoonleft  (0,1) = (1,1) $
\end{itemize}

\noindent The particulars of their derivation depend on their axioms for $\upharpoonleft$, whose presentation would take us too far afield.

At first glance this may seem like a formal novelty, but it actually hints at some very deep aspects.   Kauffman \cite{KauffmannSS}, following Spencer-Brown \cite{SpencerBrown1967}, has shown that a great variety of phenomena in philosophy, phenomenology, physics and psychology can be built up from this demi-negation operator (often referred to as a "distinction" operator).   One could take a different approach to the subject matter of this paper, beginning from distinctions of this nature and gradually building up the various complexities of logics, programs and so forth on this foundation.   Such an approach might have significant advantages from a cognitive science perspective, and in terms of foundational conceptual clarity, but would have the disadvantage of making less clear the numerous connections with ideas and results from existing computer science and mathematics literatures.

\subsection{PLN and Probabilized Constructible Duality Logic}

We now show that probabilizing constructible duality logic yields essentially the form of probabilistic logic described in \cite{PLN} as "Probabilistic Logic Networks with simple truth values."    

PLN as an overall framework has many different aspects, including e.g.

\begin{itemize}
\item various heuristic formulas for estimating the results of different inferences given inadequate information and reasonable heuristic assumptions
\item multiple sorts of truth values representing higher-order probability distributions to various degrees of approximation
\item approaches to cognitively relevant special forms of inference like intensional, causal and temporal inference
\end{itemize}

\noindent What we will explicitly address here is simply the correspondence between CD's basic approach to truth values, and the simplest of PLN's approaches to truth values.  We believe the more complex aspects of PLN can also be effectively represented in terms of the PLN/CD correspondence outlined here, but will not go into details here in that regard.

\subsubsection{Mapping Between P-Bits and PLN Simple Truth Values}

The "simple truth values" (STVs) traditionally used in  PLN \cite{PLN} have two forms,

\begin{itemize}
\item  $(s,n)$ where $s$ is a probability and $n$ is the weight of evidence, i.e. the number of observations on which $s$ is based.    
\item $(s,c)$ where $c$ is a "confidence" value heuristically determined via $c = \frac{n}{n+k}$, where $k$ is a "personality parameter."
\end{itemize}

\noindent We will work here with the $(s,n)$ version.

It will be useful here to break the $(s,n)$ value down by writing 

\begin{eqnarray*}
s = \frac{n^+}{ n^+ + n^-} \\
 n = n^+ + n^-
\end{eqnarray*} 
 
 \noindent where $n^+$  and $n^-$ denote the amounts of positive and negative evidence regarding a given proposition.   This suggests an equivalent form for the STV, namely 
 
 $$
t_{\textrm{PLN}} = (n^+, n^-)
 $$ 
 
 \noindent -- which contains the same information as $(s,n)$ but begins to lead us toward the parallel between PLN and CD logic.
 
Now suppose the $n$ observations involved in estimating a certain STV are drawn from an ensemble of $N$ different situations, where each situation may involve a single positive observation, a single negative observation, neither or both.   We may call this sort of elementary situation a {\it micro-situation}.   Any situation can obviously be decomposed into a set of micro-situations.

Directly probabilizing the p-bits associated with the $N$ micro-situations in the ensemble yields 2-component truth values of the form

$$
t_{\textrm{para}} = (\frac{n^+}{N}, \frac{n^-}{N})
$$

\noindent where $n^+$ denotes the number of micro-situations in the ensemble that contain a positive observation (which is equal to the number of positive observations made across the ensemble, because the situations are micro), and $n^-$ denotes the number of micro-situations in the ensemble that contain a negative observation (which is equal to the number of positive observations made across the ensemble, because the situations are micro).

There is an obvious relationship

$$
t_{\textrm{PLN}}  = N t_{\textrm{para}}
$$

\noindent which allows us to express a PLN STV as a renormalization of a probabilized paraconsistent truth value.   

This representation of $t_{\textrm{para}}$ is convenient for the mapping to PLN, but in some ways is less elegant than the representation in Kauffmann and Collings \cite{skilling2019} in which, on the real plane, we map $(0,1)$ to Neither, $(1,0)$ to True, $(-1,0)$ to Both and $(0,-1)$ to False.   We may call these two options the CD and KC styles of p-bit representation.   There is a linear transformation between these two (a rotation by $\frac{3 \pi}{4}$ and then a shift to the left by $1$ to get from CD to KC), so that in the end the choice is mostly a matter of taste.   In the KC representation demi-negation is represented as rotation by $\frac{\pi}{2}$ and negation as rotation by $\pi$, which is simple and beautiful.   In the CD representation negation is represented by $\neg(a,b) = (1,1) - (a,b)$ and demi-negation looks like $\upharpoonleft(a,b) = (1-b,a)$.

\subsubsection{Dependencies, Superposition and Collapse Among P-Bits}

Next we explore the algebraic operations appropriate for performing conjunctions and disjunctions on these various forms of truth values -- a topic that brings us face-to-face with some subtleties regarding probabilistic dependency among p-bits, and some aspects that begin to smell intriguingly quantum-ish.

Looking at p-bits observed across an ensemble of situations, it's clear that if we have two disjoint subsets $S_1$ and $S_2$ of the ensemble and calculate the probabilized 4-valued paraconsistent truth value across each of them yielding $t_{\textrm{para}}(S_1)$ and $t_{\textrm{para}}(S_2)$, then one has

$$
t_{\textrm{para}}(S_1 \cup S_2) = t_{\textrm{para}}(S_1) + t_{\textrm{para}}(S_2)
$$

\noindent and similarly for the $t_{\textrm{PLN}}$ renormalization.  This works for both CD and KC representations due to the linear transformation between them.

The question of $t_{\textrm{para}}(S_1 \cup S_2)$ is where things get interesting.  

In PLN, using the $s_{\textrm{PLN}}=(s,n)$ formulation of a PLN STV, one has the elementary result that if the bodies of evidence $E_1$ and $E_2$ underlying the two truth values $(s_1,n_1)$ and $(s_2,n_2)$ are independent, 

$$
s_{\textrm{PLN}}(E_1 \cap E_2) = (s_1 * s_2, n_1 + n_2 - n_1*n_2/U)
$$

\noindent where $U$ is the assumed relevant universe size.

It is not obvious, however, that the conventional probabilistic notion of independence is the right way of looking at p-bits.   Conceptually, when we say $E_1$ and $E_2$ are independent, what we mean is that knowing the truth value corresponding to any one of the observations in $E_1$ doesn't give any information about the truth value of any of the observations in $E_2$.   We could formalize this by looking at possible universes, where each universe is a set of micro-situations, defined via subsampling or the other methodologies reviewed above.

But this notion becomes a little subtler when we look at a p-bit associated with a micro-situation as the atomic form of observation.

Consider, for instance, two p-bits $b_1$ and $b_2$ that are coupled in the following way: It is known that either both lie in the set $\{True, False\}$, or both lie in the set $\{Both , Neither\}$.  Conceptually, if we think of $\{True, False\}$ as "collapsed" truth values and $\{Both , Neither\}$ as "superposed" truth values, then: Either $b_1$ and $b_2$ are both collapsed or they are both superposed.   Suppose that, across the relevant ensemble of possible universes, the both-collapsed and both-superposed outcomes occur equally often; and that within these options, True vs. False and Both vs. Neither occur equally often.

In this case, the positive bits in the two p-bits are going to be probabilistically independent of each other, as will the negative bits.   However, the p-bits are obviously dependent: Knowing whether one of them is collapsed vs. superposed tells you whether the other one is, which narrows down by half the possible states the other one may be in.

The conclusion is that, where p-bits are concerned, we need to think about dependency in a 2D rather than 1D sense.   We should consider $b_1$ and $b_2$ as independent only if knowing about the 2 components of $b_1$, taken as a set, gives you no information about the 2 components of $b_2$ (and vice versa).   In the PLN STV case, we can get away with thinking about dependency in a 1D sense, but in the p-bit case we are adding an additional degree of freedom (represented e.g. by the multiplier $N$ used in the translation between PLN STV and p-bit forms), so where dependency (and hence conjunction) are concerned we are more fully and irrevocably in the 2D world.

\section{Dependent Types for Probabilistic and Metagraph Based AI}
\label{sec:types-background}

In this section we veer in a somewhat different direction and provide some preparatory setup, explaining how dependent type theory can be utilized to manage types representing various forms of uncertainty.   These representations will then be used in Section \ref{sec:types} to provide a programmatic, type-theoretic analogue to the logical formalisms from Section \ref{sec:CD-logic}.

Constructive type theory, otherwise known as Martin-Lof type theory or intuitionistic type theory, is a variety of dependent type theory that forms a natural match for AGI in that a key theme of AGI is the need for systems to build up their core knowledge from experience and ground their abstractions in experience, and the core concept of intuitionism/constructivism is that structures need to be grounded, i.e. built up from direct examples and observations.

Here we explain how to extend standard constructive type theory to handle the probabilistic and metagraph based types that are needed for conveniently expressing the broad palette of AI algorithms used in OpenCog and other modern hybrid AI systems.  These ideas will make it fairly clear how to encompass CD and PLN within constructive type theory as well.

\subsection{Constructive Type Theory}

Roughly speaking, constructive type theory can be considered as a different way of extending first-order logic, with some similarities to and overlaps with higher-order logic and also some advantages.  

First-order logic deals with logical propositions about simple logical atoms (which may represent perceptions of some AI system, for instance, or else just basic assumed mathematical structures).  Higher-order logic accepts relations among propositions as first-class citizens -- one can can quantify over relations, and over relations of relations, etc.  

We can think of higher order logic as first-order logic equipped with a way of introducing new domains of quantification, defined to incorporate all the relationships among sets of already-existing domains of quantification.   Constructive type theory \cite{backhouse1989constructive} gives additional and more flexible options for introducing new domains of quantification.  

In constructive type theory we can introduce unspecified family symbols. We can introduce, e.g., a  a family of types  $\tau$ over the individual domain $S$, i.e.

$$
\tau(x) \textrm{ type } (x:S) .
$$

\noindent If $S$ is mapped into some interpretation $A$,  $\tau$ can then be interpreted as any family of sets indexed by $A$.   Various operations are then introduced for producing new domains of quantification based on old ones and using type families, e.g. the dependent product $\Pi$ and dependent sum $\Sigma$.  

One can build a natural interpretation of constructive logic by interpreting propositions as constructive types.   Modern dependent type based programming languages like Idris and Agda are  based on variants of this sort of constructive type theory and associated interpretation.

The theory of constructive types leads in a number of interesting advanced directions, many of which center on identity types -- types defined as families of proofs that $a=a$ -- which can be handled in various sophisticated ways.   If identity types are considered as distinct from propositions (which makes sense in terms of the use of constructive types to guide practical programming language design), then one is led into intensional constructive type theory and toward homotopy type theory.   One then obtains a variety of mathematically and conceptually fascinating results, such as theorems embodying variants of the axiom of choice \cite{program2013homotopy}, extending the already-interesting treatment of the axiom of choice in constructive type theory.    We will not need to explicitly consider these subtleties here, but we mention them as it seems possible they may be relevant to the AI directions motivating the current work.

\subsection{Probabilistic Dependent Types}

A number of approaches have been tried in order to add probabilistic types to standard type theory.   We find the approach taken in \cite{warrell2016probabilistic} especially simple and appealing, and conceptually straightforward to extend to include additional probabilistic types of interest for advanced AI.   

The basic concept is to augment standard constructive type theory with a new primitive $\textrm{random}_\rho $ which denotes sampling from a Bernoulli distribution, and which comes along with some (obvious) rules for probabilistically-weighted beta reduction on pseudo-expressions in the dependent type language \footnote{These are referred to as pseudo-expressions because they are not necessarily type-correct; the reduction rule can be applied even if the "type pseudo-expression" $\tau$ is not type-correct}.

\begin{eqnarray*}
\textrm{random}_\rho ( \tau) \rightarrow^\rho_\beta ( \tau \textrm{ true}) \\
\textrm{random}_\rho ( \tau) \rightarrow^{1-\rho}_\beta ( \tau \textrm{ false}) 
\end{eqnarray*}

\noindent which has the meaning that the type of the expression $\textrm{random}_\rho ( \tau)$ inherits from $\tau$ with with probability $\rho$ and inherits from $~\tau$ with probability $1 - \rho$.

Introducing the   $\textrm{random}_\rho $ primitive into dependent type theory allows one to create complex dependent types that reduce to a certain type with a certain probability -- different from the usual case where the question of whether a certain reduction is valid necessarily has a definite yes or no answer.

\subsubsection{Observation of Probabilistic Dependent Types}

Unlike most probabilistic programming frameworks, Warrell's approach in \cite{warrell2016probabilistic} does not include an $\textrm{ observe}$ statement that allows one to condition distributions on specific observations.   Some probabilistic programming languages separate observation from inference syntactically, others pack the two into a single two-argument "observe" statement \cite{wood2018-probprog}, which in the current context and notation could e.g. take the form

$$
\textrm{ observe } ( \textrm{random}_\rho (\tau) | \tau_1 : \tau_2)
$$

\noindent where $\tau_1 : \tau_2$ is an observation of a verified type inheritance.

For instance, suppose $\tau(n)$ and $\tau_0(n)$ denote types inheriting from the dependent type $\sigma(n)$ that is inhabited by integers that are $n$ or greater.  Suppose 

$$
\tau_1 = random_\rho( \tau) + random_\rho (\tau_0)
$$

\noindent, and $\tau_2(n)$ is a type inhabited by integers that are $>2n$.   

Now we will want to construct a probability for $\tau$ conditioned on the statement $\tau_1 : \tau_2$.   Let $\omega$ denote the percentage of the time that, when looking at a situation involving reducing $\tau_1$ (via sampling) and obtaining a result that is a type inhabited by integers $>2n$, one is looking at a situation where $ \textrm{random}_\rho (\tau)$ reduces to $\tau$.

(Quantitatively, even if $\rho$ is quite small, in this case $\omega$ is going to be $> \frac{1}{2}$ because the fact that $\tau_1$ yields an integer $>2n$ makes it necessarily the case that one of the reductions $\textrm{random}_\rho (\tau)$ and $\textrm{random}_\rho (\tau_0)$ came out positive, and potentially the case that both did.)

Then the semantics here is that

\begin{eqnarray*}
 \textrm{ observe } ( \textrm{random}_\rho (\tau) | \tau_1 : \tau_2) \rightarrow^\omega_\beta ( \tau \textrm{ true}) \\
 \textrm{ observe } ( \textrm{random}_\rho (\tau) | \tau_1 : \tau_2) \rightarrow^{1-\omega}_\beta ( \tau \textrm{ false}) 
\end{eqnarray*}

Warrell argues that  $\textrm{ observe}$ doesn't need to be taken as a primitive because the same effect can be achieved via constructing probabilistic functions with the semantics of a constraint like  $\tau_1 : \tau_2$.   Removing the $\textrm{ observe}$ primitive does simplify some formal analyses, however for our purposes it seems more convenient to act as if $\textrm{ observe}$ is there, whether as a primitive or derived construct.

\subsubsection{Comparison to Anglican}

As a stupidity-check to make sure we are not venturing too far into an idiosyncratic interpretation of probabilistic programming, let's briefly compare this to what happens in a real-world PPL -- e.g. let's consider e.g. Wood's Anglican probabilistic programming language \cite{tolpin2016design} which involves three primitives of the form

\begin{verbatim}
[ assume symbol < expr >]
[ observe < expr > < const >]
[ predict < expr >]
\end{verbatim}
 
In our considerations here so far, the distribution assumed is a Bernoulli distribution with probability $\rho$, and the reduction of a $\textrm{ random}_\rho$ pseudo-expression is what in Anglican would express as an $\textrm{ observe}$ enacted on a Bernoulli distribution with no constraint.   The $\textrm{ predict}$ action in Anglican essentially outputs the results of the various inferences made via enacting  $\textrm{ observe}$ on a distribution, which constitutes the system's knowledge about the distribution.   To some extent, new probabilistic inferences can then be made via combining these historical results, without doing further samples.

\subsection{Probabilistic Dependent Types with Sampling from Distributions over Metagraphs}

To extend the above framework for dependent type based probabilistic programming to an OpenCog-type AI context, one needs to expand beyond the Bernoulli distribution -- and also beyond distributions over simple discrete or continuous spaces to distributions over the most directly relevant spaces, e.g. graphs, hypergraphs and metagraphs.

There is a substantial literature describing probability distributions over graphs, e.g. \cite{holland1981exponential}.   Suppose we want to pick, say, a metapath $m$ within a metagraph $M$ according to some distribution $\nu$ over the space of metapaths in $M$; and suppose we have a dependent type $\tau(m)$ that takes a metapath $m$ as an argument.    We can then look at a parametrized beta-reduction such as

$$
\textrm{random}_\nu ( \tau)  \rightarrow^\nu_\beta ( \tau(m) \textrm{ true})
$$

\noindent which reduces $\textrm{random}_\nu ( \tau )$ into $ \tau(m) $ for some $m$ selected from $\nu$.  

It may be interesting to carry out this sort of reduction for multiple different distributions $\nu$ in coordination; for instance, looking ahead to Section \ref{sec:sorites}, one may look at a distribution $\nu^+$ associated with positive evidence for some statement, and another distribution $\nu^-$ associated with negative evidence for that statement.

\subsection{Dependent Types with Higher-Order or Imprecise Probabilities}

One could also extend this approach to probabilistic dependent types to higher-order probabilities in various ways.  For instance consider an imprecise probability distribution \cite{walley2000} that is characterized, not by a single parameter $\rho$, but by upper and lower bounds $(\rho_L, \rho_U)$ which connote the boundaries of an envelope contain the means of various first-order probability distributions (which could be Bernoulli distributions).   The distribution of means over the interval $(\rho_L, \rho_U)$ is then most simply assumed to be a beta distribution.

In this case one would have

$$
\textrm{random}_{\rho_L, \rho_U} ( \tau)  \rightarrow^{\rho_L, \rho_U}_\beta ( \textrm{random}_\rho ( \tau) )
$$

\noindent where $\rho$ is chosen from the interval $(\rho_L, \rho_U)$  according to the assumed distribution over this interval.

The same approach may be extended to handle more richly parametrized second order or higher order probability distributions, or more refined variants of imprecise probabilities such as indefinite probabilities \cite{PLN}.

\subsection{Gradual Dependent Types}

Gradual typing, intuitively, refers to programming languages in which the assignment of definite types to expressions in a program is optional -- and the precision of the types associated with particular expressions may change during runtime.   

In a gradually typed program, some expressions may get explicitly defined types, others may get no type assignments, and some may get higher level type assignment but not precise ones (e.g. a variable could be assigned as a Number, without specifying whether it's an Int or a Float).   Type inference as program execution progresses may then narrow down what types a certain expression is logically required to have -- and this may be contingent on various things including external input to the program.  

Among the various formalizations of the gradual typing concept, the approach from \cite{garcia2016abstracting} is especially direct and simple.   Here the gradual type of each expression, at each point in a program's execution, is considered as a set of definite types.  There is an extreme unspecified case where an expression is associated with the set of all possible definite types in the language, and an extreme specified case where a maximally precise lowest-level type is associated, and then there are intermediate cases.   At each point in the execution of a program, an expression will have certain constraints on what types it may have, based on its relationships with other expressions, which narrows down its gradual type set.  The process of program execution then involves ongoingly solving the constraint satisfaction problem of mutual type constraints.

The intersection of gradual and probabilistic dependent types appears straightforward formally.   What is most interesting here conceptually is the possibility that when a statement does not have a definitely known type, it may still have a probabilistically known type.  So in a probabilistic typing framework, certainly known types, totally unknown types and types restricted to a certain higher-level type but not known in detail -- all appear as special cases of the association of an expression with a probability distribution over type space.   The gradual typing of expressions during the course of running a program is then best framed as a gradual decrease in the entropy of the distribution over type space associated with expressions.

The general framework one arrives at is: An expression in a language has a set of types associated with it, at varying levels of abstraction, and each one labeled with a certain probability.    In the simplest case one can view one of the jobs of the program execution engine as maintaining consistency among the probabilities assigned to the different types for each expression.   Allowing inconsistencies here is also feasible but complicates things significantly, a topic that relates to our concerns in section \ref{sec:sorites} below.   In this "paraconsistent" case, carefully accounting for and managing inconsistencies becomes critical for efficient program execution in practical cases.

\section{The Programming Language Cognate of Constructive Duality Logic}
\label{sec:types}

The next question to consider is what analogue one obtains in the programming language domain by applying the Curry-Howard correspondence to Constructive Duality logic.

There is a well known mapping between constructive type theory, constructive logic and typed lambda calculus \cite{sorensen2006lectures}.  Extending the logic/lambda portion of this mapping, Patterson \cite{patterson1998implicit} presents a Curry-Howard mapping of CD logic into a variant of lambda calculus called $\lambda_{\textrm{HAZ}}$.   If one removes negative content, projecting to positive truth values only, then one gets a standard Curry-Howard mapping from intuitionistic logic into lambda calculus or type theory.   Similarly if one removes positive content, projecting to negative truth values only, then one gets a dual Curry-Howard mapping from intuitionistic logic into lambda calculus or type theory.   The $\lambda_{\textrm{HAZ}}$ calculus embeds both of these mappings.

The type-theoretic analogue to CD logic has not been fleshed out as thoroughly in prior literature -- so we will make an effort in this direction here, with the caveat that we have not yet attempted to work out all the formal details.  Starting with the well known mapping from constructive logic into constructive type theory, it would appear we can immediately map CD logic into a programming language in which a program comprises a set of dependent type expressions, each one associated with two different execution traces, which we may call a {\it positive trace} and a {\it negative trace}, defined as follows:

\begin{itemize}
\item The positive trace, if not empty, comprises a series of steps equivalent to the construction of a proof that the type is inhabited.   
\item The negative trace, if not empty, comprises a series of steps equivalent to the construction of a proof that the type is empty (i.e. that inhabiting the type implies inhabiting the empty type).
\end{itemize}

\noindent Table \ref{tab:traces} concisely summarizes the logical and program semantics involved here.
\begin{table}[ht]
\caption{Logical and Procedural Semantics of Positive / Negative Traces} 
\centering 
\begin{tabular}{c c c c} 
\hline\hline 
Positive Trace & Negative Trace &  Logical Semantics & Program Semantics \\ [0.5ex] 
\hline 
Nonempty & Empty & True & Typed Lambda Calculus \\ 
Empty & Nonempty & False & Typed Lambda Calculus \\ 
Empty & Empty & Neither True nor False & Null \\ 
Nonempty & Nonempty & Both True and False & $\lambda_{\textrm{HAZ}}$ calculus\\ 
\hline 
\end{tabular}
\label{tab:traces} 
\end{table}

\subsection{Paraconsistent Logic and Gradual Typing }

Gradual type systems can be naturally mapped into paraconsistent logic.   This has been observed informally before, but we are not currently aware of prior research papers treating the correspondence in a rigorous way.

To illustrate the points involved here, suppose that in a gradual typing system one has a type $T$ with subtypes $T_1$  and $T_2$.   Suppose $T_1 \land T_5$ is nonempty whereas $T_2 \land T_5$ is empty.   Then we can say that where $x:T$, $T \land T_5$ is both true and false.   Proof that $x$ is $T_1$ is the positive trace, proof that $x$ is $T_2$ is the negative trace.  If $T_7$ is a subtype of $T$, then there may be both positive, negative and overdetermined options (i.e. $T_7$ could inherit from both $T_1$ and $T_2$, either or neither).

Conceptually, if there is the possibility that $T$ may never be resolved to $T_1$ and $T_2$, then paraconsistent logic is the correct model of the gradual type system.

In the simplest implementation of gradual-typing based programming languages, program interpretation is restricted to cases corresponding to truth value $(1,0)$.   In the above example, that would mean if the interpreter knows $x:T$ but doesn't know whether $x:T_1$ or $x:T_2$, then it would just avoid interpreting anything regarding $x \land y$ where $y:T_5$.   

This simple mode of interpretation is reasonable in many contexts but not necessarily in an AGI or real-time-control-oriented narrow-AI context, where we may not know whether the choice $x:T_1$ or $x:T_2$ is ever going to get resolved, and even if we have reason to believe it's going to get resolved eventually, we may need to make a practical judgment before it is.   So in these cases, a more aggressive mode of interpretation may be needed, wherein the interpreter explores the two options $x:T_1$ or $x:T_2$ and keeps track of the sometimes-conflicting conclusions that may occur in each case.   This leads directly to paraconsistent logic and probabilistic reasoning; and when a system then needs to make a practical choice of action based on its interpretation process, it needs to carry out a programmatic action that is isomorphic to the PLN Rule of Choice.

\subsection{Negation and Continuation in Intuitionistic and Classical Logic }

When considering the implications of these ideas for practical programming language design, the connection between negation in intuitionistic logic and continuation in programming languages becomes relevant.  This connection has been fleshed out in detail from a number of perspectives \cite{reynolds1993discoveries} and here we will only scratch the surface.   

A continuation, in the programming language context, is an abstract representation of the control state of a computer program.   The term "continuation" is also used to refer to constructs that give a programming language the ability to save the execution state at any point and return to that point at a later point in the program, possibly multiple times.   

In  continuation-passing style (CPS) programming, a species of functional programming,  control is passed explicitly from one function to another in the form of a continuation.  A function written in continuation-passing style takes an extra argument: an explicit "continuation", which is a function of one argument.  When the CPS function has computed its result value, it "returns" it by calling the continuation function with this value as the argument. This means that when invoking a CPS function, the calling function is required to supply a procedure to be invoked with the subroutine's "return" value.  

To connect this with logic, recall that in intuitionistic logic, we can think of $\neg \tau$ as being equivalent to $\tau \rightarrow \textrm{False}$, or, as the type $\tau \rightarrow \bot$, where $\bot$ is any uninhabited type.  That is, if $\neg \tau$ is true, then if you give me a proof of $\tau$, I can give you a proof of $\textrm{False}$.

So, suppose that we have a special type $\mathcal{A}$ that is the return type of continuations. That is, a continuation that takes an argument of type $\tau$ has the type $\tau \rightarrow \mathcal{A}$.   Assume further that we have no values
of type $\mathcal{A}$, i.e., $\mathcal{A}$ is an uninhabited type -- because continuations, by design, don't return anything.

A continuation-passing style translation of an expression $e$ of type $\tau$ , $CPS[[e]]$, has the form $\lambda k : \tau \rightarrow \mathcal{A}$ , where $k$ is a continuation, and the translation will evaluate $e$, and give the result to $k$. Thus, the type of $CPS[[e]]$ is $(\tau \rightarrow \mathcal{A}) \rightarrow \mathcal{A}$.  Under the Curry-Howard isomorphism, this type corresponds to $(\tau \rightarrow False) \rightarrow False$, or, equivalently, $\neg(\neg\tau )$, the double negation of $\tau$ , which is equivalent to $\tau$.  CPS translation thus converts an expression of type $\tau$ to an expression of type $(\tau \rightarrow \mathcal{A}) \rightarrow \mathcal{A}$, which is equivalent to $\tau$.

Notably, the computational behaviors of $A \rightarrow \bot$ and $\neg A$ may be different even though the two are logically equivalent.   A closed function of the form $A \rightarrow \bot$  is dead code that can never be called, whereas . a continuation of type $\neg A$ can conceivably be thrown to: it provides an "escape hatch" if one does encounter an $A$ but does not guarantee no closed terms of type $A$ will ever be found.

The symmetric lambda calculus \cite{filinski1989declarative} provides a theoretical view that clarifies such complexities, representing functions and continuations as symmetrical, dual aspects of a richer sort of lambda calculus that explicitly represents both.   Crudely put, a function represents the path from the beginning of a computation to a certain intermediate point; whereas a continuation represents the path from a certain intermediate point to the end of a computation.   The symmetric lambda calculus abstractly encapsulates the symmetry between moving forward from the beginning and moving backwards from the end.

Munch \cite{munch2014formulae} has noted some advantage to dealing with captured stack states instead of continuations in the context of Curry-Howard mappings of intuitionistic logic.   They show that the stack state, captured at a certain point in a computation, has similar algebraic properties to the continuation, and leads more directly to involutive negation.   Overall, the question of the most elegant way to represent various forms of negation on the programming language side of the Curry-Howard correspondence (be it continuations, stack state captures, or something else in the same conceptual vicinity) is an intriguing and pragmatically relevant topic needful of further investigation

It was shown long ago that one can embed classical-logic theorems in intuitionistic logic in an interesting way via strategically inserting double-negations (replacing $A$ with $\sim \sim A$, where $\sim$ is intuitionistic negation).   For example, while $A \lor \sim A$ is not provable in constructive logic, the weaker $\sim \sim (A \lor \sim A)$ is.   This  "Godel-Gentzen double-negation translation" is useful in functional language compilation, where the correspondence between $\sim$ operations and continuations can be directly leveraged.   

Note that the Godel-Gentzen double-negation translation is embeddable in CD logic since the exponential $!$ maps CD logic into intuitionistic logic, where the complement operator in the first Heyting algebra of the CD pair $\mathcal{H} \times \mathcal{H}^{op}$ is thus the intuitionistic negation $\sim$.   Dual to this translation in the "positive logic" of CD logic, there is also an embedding based on the exponential $?$, which works with the falsity-oriented intuitionistic logic  or "negative logic" corresponding to $\mathcal{H}^{op}$.  One then reconstructs dual classical logics within the two sides of CD logic.

There is an interesting connection between the function/continuation duality and the positive/negative duality in CD logic, in that the proofs corresponding to the negative trace of type $T$, would seem to have the same type as proofs corresponding to continuations of functions of type $T$.   In fact the proofs corresponding to the negative trace would seem to correspond to continuations of functions of type $T$ -- specifically, continuations that take the outputs of such functions and map them ultimately into $\textrm{False}$.   The implications of this parallel for programming language design are not obvious but seem worthy of reflection.

Continuations can also be used to enable large-scale computational operations to be broken down into manageable pieces -- e.g. \cite{Raab2016} presents a version of the catamorphism (fold) operation that performs a fold one-step-at-a-time rather than all-at-once, via using a continuation to pass state from one step of the fold process to the next.   It seems that a stack state capture would also suffice here.

\section{Probabilistic Paraconsistent Logic on Metagraphs}
\label{sec:metagraph-logic}

The CD logic framework allows us to create a four-valued intuitionistic logic from any pair of Heyting algebras.   A particularly interesting case, from an AI perspective, is the Heyting algebra of subgraphs of a metagraph.   

In \cite{Goertzel2020metagraph} we have given one way to define a Heyting algebra $\mathcal{H}$ on a metagraph $M$ -- basically, via defining a topology on $M$ and using the standard route from topologies to Heyting algebras.   On a directed or undirected typed metagraph, the appropriately defined metapaths may be taken as the open sets; especial attention is paid to the case of directed typed metagraphs (DTMGs). 

The opposite algebra $\mathcal{H}^{op}$ is then defined as the algebra with the same elements but the opposite ordering of $\mathcal{H}$.    We can then use $\mathcal{H} \times \mathcal{H}^{op}$ to construct a CD logic on a metagraph.

Given a probability distribution over a space of typed metagraphs built with a common type system, one can then construct a PLN logic on a metagraph, using the probabilization method outlined above.

\subsection{Using Intuitionistic Probabilities For Averaging Over Possible Situations}

In the treatment above, we have implicitly assumed that the probabilities used for averaging over possible situations are ordinary set-based probabilities.   However, it is relevant to recall here that, as shown in \cite{Goertzel17a}, from any Heyting algebra one can derive a sort of "intuitionistic probability" that obeys all the standard symmetries of probability theory (but does not obey excluded middle).    \footnote{In \cite{Goertzel17a} a specific Heyting algebra is defined on graphs/ hypergraphs / metagraphs, which I now believe is generally going to be less useful for AI than the Heyting algebra derived from metapath topology on metagraphs.   However, the point from \cite{Goertzel17a} that Heyting algebras lead to intuitionistic probability distributions remains relevant regardless of how the Heyting algebra on metagraphs is set up.}

If we assume that the "situations" over which averaging is done are represented by metapaths within a metagraph, then the process of averaging over situations is naturally done in an intuitionistic probability setting, where the Heyting algebra is that of metapaths.   This is what one is dealing with in a reflexive framework where the (experienced and hypothetical) situations used for assessing the truth values of propositions are considered as part of the same knowledge graph as the propositions themselves.

\subsection{Semantics of CD/PLN Logics Grounded in Metagraphs}

But what is the semantics of the PLN logic grounded in a metagraph in this way?

The subgraphs of the metagraph are being taken as the primitive, atomic observations underlying the PLN semantics.   In the case where these subgraphs represent direct perceptions of some external world, or conjectured perceptions of an external world, one then obtains a logic that builds up semantics via counting occurrences of percept-representing subgraphs within possible-situation-representing subgraphs.    In the case where these subgraphs represent abstract knowledge (or machinery used for manipulating abstract knowledge, which may be considered a sort of abstract knowledge), one has a logic of meta-reflection,  i.e. reasoning about the internal structures and processes of a knowledge-manipulating and -creating system.

\subsection{Homotopy Type Theory in Metagraph-Based Logic Systems}

If one has a logic defined via Heyting algebras of subgraphs of a DTMG, and also embeds the inference rules of the logic system within the metagraph, then one has a situation where a chain of inference steps leading from premise-set $A$ to conclusion $B$ forms a metapath through the metagraph.   One can form a "logical topology" over the space defined as the union of all such "logical metapaths" within a given metagraph by considering the logical metapaths as open sets.

In the context of this topology, it's interesting to look at transformations that are smooth in some sense but are not necessarily homomorphisms.  For instance, if one has a proposition of the form $P_1(x) =$ "$x$ connects to $S_1$" and a similarly defined $P_2$, one can define

\begin{mydef}
$P_1 \rightarrow_{\textrm{sm}} P_2$ if and only if there is a smooth transformation taking $P_1$ into $P_2$
\end{mydef}

In the logical topology, a homotopy class of paths from $A$ to $B$ is precisely an equivalence class of proofs deriving $B$ from $A$, with the property that any proof in the class can be "deformed" into another proof in the class by a series of elementary smooth transformations.

\subsection{Probabilistic Programming for Inference Control}

We can also straightforwardly frame history-based inference control as described in \cite{EGI2} and prototyped in the current OpenCog system in paraconsistent-logic terms.   

Here one considers an inference engine making choices of what inference steps to follow by considering its current inference context -- where an inference context comprises the collection of inference steps that has been taken so far in an in-progress inference process, combined with other information about the reason and setting in which that inference process has been undertaken.   An inference context may include some hypothetical assumptions made in the course of inference (in the case of standard backward chaining inference these hypotheses will often be iteratively nested).   Relative to an inference context $\mathcal{C}$, one is interested to assess the probability of a certain new inference step (or collection of inference steps) $\mathcal{I}$, if executed and added onto the previous ones in the context, leading usefully toward a relevant inference goal (e.g. deriving a given conclusion, or creating new hypothesis with high surprisingness value, etc.).    

Given a specific hypothetical situation $S$ consistent with  $\mathcal{C}$, the inference engine can estimate $\nu_S(\mathcal{I})$ -- i.e. whether or not $\mathcal{I}$ is useful as a next inference step, in the case that $S$ holds.   In a paraconsistent context, there may be evidence that this is the case and also evidence that it's not, both of which contribute to components of $\nu$.   Summing up these truth values over various situations and then normalizing one gets a PLN truth value for the odds that $\mathcal{I}$ will be useful given $\mathcal{C}$.

\section{Probabilistic Paraconsistent Concept Boundaries}
\label{sec:sorites}

Now we connect the various threads from the preceding sections, fleshing out more thoroughly how paraconsistent logic and PLN probabilistic logic map into the generalized form of probabilistic programming outlined in Sections \ref{sec:types-background} and \ref{sec:types}.

To make the connections clear, we will frame the discussion in terms of a particular practical example of paraconsistent reasoning, drawn from Weber's paraconsistent analysis of vagueness in the context of the {\it sorites} paradox.   However, the ideas presented are not restricted to this example.

The sorites paradox, which gets its name from the Greek word {\it soros} for "heap", is classically framed in terms of the question "How many grains of wheat does it take to make a heap?"  It seems that no single grain of wheat can make the difference between a number of grains that does, and a number that does not, make a heap.   One can then reason as follows: Since one grain of wheat does not make a heap, it follows that two grains do not; and if two do not, then three do not; and so on. The conclusion is as clear as it is absurd: A million grains of what do not make a heap.   No number of grains of wheat make a heap.

This sort of confusion about concept boundaries actually does play a role in everyday life -- eating one more chocolate won't make much difference to my weight or health, right?   Where exactly is the boundary between a democratic and authoritarian society?   Can we really say that just a tiny bit less democratic control really makes the critical difference?

One can view the sorites "paradox" as an illustration of the problem of trying to represent and analyze fuzzy / probabilistic concepts using crisp models.  On the other hand, we do habitually model all these fuzzy/probabilistic concepts using crisp models, terms and symbols -- and this sort of crisp reasoning about fuzzy concept boundaries does have the hallmarks of paraconsistency.   That is, there is a certain sometimes-confusing inconsistency involved, which sometimes does lead to strange conclusions.   However, we are able to navigate around and successfully work with this inconsistency most of the time.   The inconsistency regarding the boundaries of any particular concept doesn't generally pollute our whole minds and make everything we think inconsistent, though it may impact some of our practical judgments about that concept and closely related things.

We will argue that the sorites paradox is not just a curiosity, but actually highlights significant issues regarding the fuzzy boundaries of most natural concepts.   Considering the relation between paraconsistent and probabilistic logic in the context of the sorites paradox yields generally meaningful insights regarding the architecture of conceptual ambiguity.   This is much in the spirit of conceptual/mathematical insights that have been found via following other classical paradoxes -- Godel's Theorem's "This statement is not provable" following Epiminides' paradox "This sentence is false", Chaitin's algorithmic-information formulation of Godel's Theorem following Berry's paradox "The first number that can be described in one hundred words or less", and so forth.

Weber \cite{weber2010paraconsistent} gives a formal analysis of the sorites-paradox-type "concept boundary" situation using paraconsistent logic.   The specific version of paraconsistent logic used is different from the CD version we've been pursuing here, but this difference is not critical to his analysis.   

\begin{figure}
\centering
  \includegraphics[width=7cm]{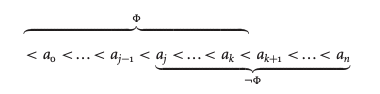}
  \caption{General depiction of the {\it high-up} predicate for number series.  The high-up portion of the series intersects with the not-high-up portion of the series in the cutoff region of the series (the latter being amenable to modeling using paraconsistent, overdetermined truth values).}
  \label{fig:high-up1}
\end{figure}

\begin{figure}
\centering
  \includegraphics[width=5cm]{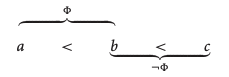}
  \caption{A simple model of  the {\it high-up} predicate for number series, stripped down to the basics for maximally concise formal analysis. }
  \label{fig:high-up2}
\end{figure}

He analyzes a simple toy situation involving the predicate {\it high-up} among numbers, and presents an analysis that clearly generalizes beyond this case.   Figure \ref{fig:high-up1} presents the  general schematic considered for  {\it high-up} -- a series of numerical values, some of which are high-up, some of which are  not high-up, and some of which are on the boundary.   Figure \ref{fig:high-up2} depicts the simplest possible case of this schematic, which is the one Weber analyzes in detail.  Once one understands the simple case it's clear that the more general case, and other instances of fuzzy concept boundaries, can be handled the same way.

\subsection{A Paraconsistent Logic Model of the Sorites Paradox}

To formalize the simple case, consider the predicate $\Psi = high-up$, and consider the following predicate indicating what it means for a certain value $z$ to be a "cutoff" or boundary value of a predicate such as $\Psi$:

$$
\textrm{cutoff}(\Psi,z) = \neg \Psi z \land (y < z \rightarrow \Psi y)
$$

Letting $\nu(z)$ denote the p-bit corresponding to $z$, consider then the following  model of the situation in Figure \ref{fig;high-up2}:

\begin{itemize}
\item  $\nu(\Psi a) = (1,0)$
\item $\nu(\Psi b) = (1,1)$
\item $\nu(\Psi c) = (1,1)$
\item $\nu(a<b) = (1,0)$	
\item $\nu(b<c) = (1,0)$	
\item $\nu(c<c) = (0,1)$	
\end{itemize}

\noindent (this is following \cite{weber2010paraconsistent} closely but changing the truth-value notation and scheme).

Note that we have

$$
\nu(\Psi b) = \nu(\neg \Psi b) = (1,1)
$$

\noindent and similarly for $c$.   We have the conclusion that both $b$ and $c$ are cutoffs; and the conclusion that: The statement that there is a cutoff for $\Psi$ is both true and false.

If we formulate

\begin{itemize}
\item $P = \exists z : \textrm{cutoff}(\Psi,z)$
\item $N = \neg \exists z : \textrm{cutoff}(\Psi,z)$
\end{itemize}

\noindent then we conclude that

\begin{itemize}
\item $\nu(P) = (1,1)$ 
\item $\nu(N)=(1,1)$
\end{itemize}

\subsection{A Paraconsistent Type-Theoretic Model of the Sorites Paradox}

To build a type-theoretic image of the above paraconsistent logic model of the Sorites paradox, consider the dependent type 

$$
\tau(x) = \textrm{cutoff}(\Psi,x)
$$

We then can see that

\begin{itemize}
\item {\bf Positive trace}: from the above model, constructs $b$ so that $\textrm{cutoff}(\Psi,b)$, and returns $b$
\begin{itemize}
\item This is a program that begins with data about the series and outputs $b$, along with an indication that $b$ is indeed a cutoff for $\Psi$.
\end{itemize}
\item {\bf Negative trace}: from the above model, calculates that $\neg \textrm{cutoff}(\Psi,b)$, $\neg \textrm{cutoff}(\Psi,c)$ and $\neg \textrm{cutoff}(\Psi,x)$ for any $x>b$ or $x>c$.  Similar calculations apply for any assignation of truth values to $\nu(\Psi b)$, $\nu(\Psi c)$.  Also proves that for any model finding a cutoff, it would need to agree with one of these assignations of truth values on $b$ and $c$.
\begin{itemize}
\item This is a program with Subroutine 1 that begins with data about b and outputs an indication that $b$ is not a cutoff for $\Psi$, and Subroutine 2 that begins with data about any series of values and outputs the output value of Subroutine 1.
\end{itemize}
\end{itemize}

Connecting this example with our earlier discussion of continuations, we can see that a continuation of the positive trace would take $b$ as an input, and would then show that actually $b$ is not a cutoff.   And generally, the two traces can be viewed as continuations of each other.  In case $(1,0)$, the negative trace is a trivial "abort" continuation; and in case $(0,1)$ the positive trace is of this sort.   In case $(1,1)$ neither is trivial.

Using a similar approach to what we've done here for {\it high-up}, we could similarly obtain a paraconsistent logical and type-theoretic model regarding the boundaries of very many natural concepts.   E.g. the boundary between man and woman,  fat and thin,  true and false in the commonsense interpretation,  cup and bowl, etc.

\subsection{From Paraconsistent Concept Boundaries to Probabilistic/Fuzzy Concept Boundaries}

Following the mapping between CD logic and PLN logic outlined above, we can take a predicate such as $\Psi$ ({\it high-up}) and consider it to vary across different situations $S$.   So in situation $S_1$ one has a situation-biased interpretation of {\it high-up} denoted $\Psi_{S_1}$, and so forth.   

As a real-world example, consider the notion of high temperature in human environments on the Earth's surface.   A high temperature in winter means one thing, a high temperature in summer means another thing.   A high temperature after a series of $45C$ days means one thing, a high temperature after a series of $40C$ days means something slightly different, etc.  If a given temperature reading has a certain paraconsistent truth value in each situation, then averaged over a number of situations, that temperature reading $z$ will end up with a two-component probabilistic truth value $t(z) = (w^+, w^-)$, which will behave in a very natural way:

\begin{itemize}
\item If the temperature $z$ is high-up in every situation, and not not-high-up in any situation, then $t(z) = (1,0)$
\item  If the temperature $z$ is not-high-up in every situation, and not high-up in any situation, then $t(z) = (0,1)$
\item  If the temperature $z$ is, in every situation, a cut-off value on the boundary between high-up and not-high-up,  then $t(z) = (.5,.5)$
\item  If the temperature $z$ is
\begin{itemize}
\item in $\frac{1}{3}$ of situations, a cut-off value on the boundary between high-up and not-high-up
\item in $\frac{1}{3}$ of situations,  high-up and not not-high-up
\item in $\frac{1}{3}$ of situations, not-high-up and not high up
\end{itemize}
\noindent then $t(z) = (.5,.5)$
\item If the temperature is 
\begin{itemize}
\item in $\frac{1}{5}$ of situations, a cut-off value on the boundary between high-up and not-high-up
\item in $\frac{3}{5}$ of situations,  high-up and not not-high-up
\item in $\frac{1}{5}$ of situations, not-high-up and not high up
\end{itemize}
\noindent then $t(z) \approx (.33,.77)$
\item If the temperature is 
\begin{itemize}
\item in $\frac{2}{5}$ of situations, a cut-off value on the boundary between high-up and not-high-up
\item in $\frac{1}{5}$ of situations,  high-up and not not-high-up
\item in $\frac{2}{5}$ of situations, not-high-up and not high up
\end{itemize}
\noindent then $t(z) \approx (.64,.36)$
\end{itemize}

\noindent. In each case, the composite strength value $s$ of $t(z)$ will tell you simply the probability with which a randomly selected piece of evidence regarding $z$ will place it on the positive side of the cutoff regarding $\Phi$.   This can also be interpreted as a fuzzy membership value in the concept $\Phi$, according to the probabilistic interpretation of fuzzy values given in \cite{Goertzel2010e}.   We thus arrive at a grounding of the fuzzy membership values associated with so-called "soggy predicates" (Simple Observationally Grounded predicates, \cite{SoggyPredicates2020}) in terms of probabilistic truth values derived from paraconsistent truth values.   

Looking at these simple examples, we see that various theoretically quite different forms of truth value (paraconsistent, probabilistic, fuzzy) are "merely" different ways of normalizing the same basic evidence-counts.   But the  "merely" here is in quotes because, in fact, each mode of normalization reflects a certain way of relating an individual value to an overall multi-situational context, leading to certain algebraic symmetries which simplify certain sorts of reasoning about individual and aggregated measurements.  Each of these sets of symmetries has practical calculational and heuristic value for cognition, which is why these various forms of truth values are all useful for intelligence, and why conversions between them are also important cognitive operations.

\subsection{Paraconsistent Concept Boundaries and Probabilistic Types}

The {\it high-up} examples also yields a clear picture of the relationship between PLN logic and a form of probabilistic programming.   Just as PLN can be obtained via averaging paraconsistent truth values over multiple situations, so a related form of probabilistic programming can be obtained via averaging the positive and negative traces (programs) associated with these paraconsistent truth values over multiple situations.

Continuing with the temperature example, suppose we randomly select a situation $S$ involving a series of temperature values from some distribution over situations.   For each $S$ selected we get 

\begin{itemize}
\item Positive trace that shows $P_S$ is inhabited by calculating $\nu( \textrm{cutoff}( \Psi , b_S) ) = (1,*)$
\item Negative trace that shows $P_S$ is not inhabited by calculating $\nu( \textrm{cutoff}( \Psi , b_S) ) = (*,1)$
\end{itemize}

To frame this in terms of probabilistic types, we would have a type of the form

$$
\textrm{random}_S 
$$

\noindent which could be used to form constructs such as

$$
random_S (b \in S \land \textrm{cutoff}(\Psi,b ) )
$$

If situations are represented as sub-metagraphs then this takes the form

$$
random_M (m \in M \land \textrm{cutoff}(\Psi,m ) )
$$

Clearly this same sort of mapping into probabilistic types works for any concept boundary, not just for high-up, and in fact for any CD-truth-valued predicate, not just for those denoting concept boundaries.

\section{Probabilistic Paraconsistent Concept Lattices}
\label{sec:concept}

The implications of these ideas for various aspects of AI reasoning, learning and memory are quite diverse; here we briefly discuss one example, which ties in closely with the above discussion of concept boundaries -- the definition of concepts themselves, built up via algebraic analysis from sets of paraconsistent properties.

Here we explore how to define the fuzzy degree to which an entity possesses some property, in a case where evidence about entity-property relationships is paraconsistent in nature.   We then build on this to suggest a theory of paraconsistent concepts that generalizes classical crisp and fuzzy Formal Concept Analysis.

\subsection{Paraconsistent Information and Intension}

An ordinary discrete probability distribution is a function that assigns a probability value to each possible outcome of some experiment -- which can be thought of as the percentage of possible situations in which that outcome is realized.   One may similarly think of a {\it paraconsistent probability distribution} as a function that assigns a value $(w^+,w^-)$ to each possible outcome $o$ of some experiment -- where $w^+ [w^-]$ can be thought of as the percentage of possible situations in which there is positive [negative] evidence in support of $o$ being the outcome.

A paraconsistent probability distribution (ppd) can be thought of as a Cartesian product of two ordinary probability distributions, i.e. $\alpha = \alpha^+ \times \alpha^-$.   The entropy of the ppd $\alpha$ then breaks down as 

$$
H( \alpha) =  H( \alpha^+) + H( \alpha^-) 
$$

\noindent and the relative entropy between two ppds $\alpha$ and $\beta$ defined in relation to the same experiment breaks down as

$$
H(\alpha | \beta) = H(alpha^+ | \beta) + H(alpha^- | \beta) 
$$

\subsection{Paraconsistent Properties}

Among other applications, this conceptualization of paraconsistent relative information provides a paraconsistent grounding for PLN's intensional reasoning \cite{PLN}, which has to do with the "pattern-sets" associated with different terms, where the degree of membership of $z$ in the pattern-set for $x$, $\Psi_{\textrm{PAT}_x}(z)$, is defined conceptually as the information content in $P(z|x)$ relative to $P(z|C)$ where $C$ is an assumed broader context.   

As an example to think about, suppose $x$ denotes whales, $z$ denotes swimmers and $C$ denotes Earth animals in general.

Suppose known examples of $x$ (known whales, in the example) are $x_1, \ldots, x_k$, and for each $x_i$ we have a paraconsistent weight pair $(x_i^+, x_i^-)$ where

\begin{itemize}
\item $x_i^+$ indicates the percentage of situations involving $x_i$ in which there is positive evidence that $x_i \rightarrow  z$ (in the example, that whale $x_i$ swims)
\item $x_i^-$ indicates the percentage of situations involving $x_i$ in which there is negative evidence that $x_i \rightarrow  z$ (in the example, this means there is negative evidence about whale $x_i$ being a swimmer)
\end{itemize}  

And if known examples of $C$ are $c_1, \ldots, c_K$ (in the example, this would be a list of known Earth animals of all sorts, whales included). then for each $c_i$ we have a paraconsistent weight pair $(c_i^+, xc_i^-)$ defined analogously.

We may summarize the above information as two distributions 

\begin{eqnarray*}
\mathcal{X}_z = \mathcal{X}_z^+ \times \mathcal{X}_z^- \\
\mathcal{C}_z = \mathcal{C}_z^+ \times \mathcal{C}_z^-
\end{eqnarray*}

\noindent, and calculate the relative entropy 

$$
H(\mathcal{X}_z | \mathcal{C}_z)
$$

\noindent This gives a measure of the extent to which $z$ lies in the intension of $x$ in context $C$, which takes into account the potentially paraconsistent nature of the truth values via which $z$ is associated with instances of $x$ and $C$.

\subsection{Paraconsistent Formal Concept Analysis}

Now how can one use paraconsistent properties to define paraconsistent concepts?   Suppose one has a set of logical terms associated with property-sets using weights defined via paraconsistent truth values -- perhaps via intensional reasoning as mentioned above or otherwise.   One can then identify concepts via variations on Formal Concept Analysis  (FCA) \cite{ganter2012formal} (what we are calling "terms" here would be "objects" in FCA lingo).   

The basic idea of crisp FCA is that, given a set $O$ of objects and a set $P$ of properties, $(O,P)$ is a formal concept precisely when:

\begin{itemize}
\item every object in $O$ has every property in $P$
\item for every object in $G$ that is not in $O$, there is some property in $P$ that the object does not have
\item for every property in $M$ that is not in $P$, there is some object in $O$ that does not have that property
\end{itemize}

Where $\mathcal{O}$ is an overall set of objects, $\mathcal{P}$ is an overall set of properties and $I$ is a binary relation specifying which (object, property) pairs are valid, the underlying mathematics of FCA is based on the Galois connection between the operators

\begin{itemize}
\item $O^* = \{p \in P | (o,p) \in I \textrm{ for all } o \in O \}$, i.e., a set of all properties shared by all objects from $O$
\item $P^! = \{o \in O | (o,p) \in I \textrm{ for all } p \in P \}$, i.e., a set of all objects sharing all properties from $P$
\end{itemize}

\noindent A formal concept is then a pair where $O^! = P$ and $P?\! = O$.

In an OpenCog context, this manifests itself as Galois connection between the operator finding all patterns that are universal in a certain object-set, and the operator finding all objects containing a given pattern.   In the current OpenCog software system, these operators are represented by DualLink and GetLink \cite{VepstasPatternMatcher2020}. 

Abner et al \cite{fuzzyFCA} and many others have explored generalizations of FCA to fuzzy property values, e.g. beginning with the Galois connection between operators

\begin{itemize}
\item $\Psi_{O^?}(a) = \inf_{o \in O} [\Psi_O(o) \rightarrow \Psi_I(o, a)]$ 
\item $\Psi_{P^!} (o) = \inf_{p \in P} [\Psi_P(a) \rightarrow \Psi_I(o,a) ]$
\end{itemize}

\noindent where $\Psi_X$ is the membership function of the fuzzy set $X$ and $\rightarrow$ is fuzzy implication.  A fuzzy formal concept is then a pair where $O^! = P$ and $P^! = O$.

We can define paraconsistent FCA similarly, given a set $\mathcal{O}$ of objects and a set $\mathcal{}P$ of properties, with $I$ defined as a mapping from $\mathcal{O} \times \mathcal{P}$ to the four CD truth values.  So we are dealing with a situation where each object-property pair $(o,p)$ may either be true, false, true and false, or neither true nor false.   In this case we would say $(O,P)$ is a formal concept precisely when:

\begin{itemize}
\item every object in $O$ is related to every property in $P$ via a truth value $(1,*)$
\item for every object in $G$ that is not in $O$, there is some property in $P$ that the object has with a truth value $(0,*)$
\item for every property in $M$ that is not in $P$, there is some object in $O$ that has that property with a truth value $(0,*)$
\end{itemize}

\noindent The relevant operators here are

\begin{itemize}
\item $O^* = \{p \in P | I(o,p) \in (1,1) \cup (1,0) \textrm{ for all } o \in O \}$
\item $P^! = \{o \in O | I(o,p) \in (0,0) \cup (0,1)  \textrm{ for all } p \in P \}$
\end{itemize}

\noindent A paraconsistent formal concept is then a pair where $O^! = P$ and $P?\! = O$.   It seems intuitively clear that the set of paraconsistent formal concepts defined in this way forms a complete lattice (as in the standard and fuzzy varieties of FCA), but this requires further exploration and formal proof.

Extending this, we can define fuzzy paraconsistent formal concepts via defining

\begin{itemize}
\item The fuzzy membership degree $\Psi_O(o) = \Psi^+_O(o),  \Psi^-_O(o))$ as the p-bit summarizing the evidence that $o$ is an instance of $O$
\item The fuzzy membership degree $\Psi_P(p) = \Psi^+_P(p),  \Psi^-_P(p))$ as the p-bit summarizing the evidence that $o$ is an instance of $O$
\item $I^+(o,p)$ and $I^-(o,p)$ as the average amounts of positive and negative evidence regarding the object-property relation $(o,p)$, calculated over an ensemble of situations.   
\end{itemize}

\noindent We can then set up the four operators

\begin{itemize}
\item $\Psi_{O^{?+}}(ap) = \inf_{o \in O} [\Psi_{O^+}(o) \rightarrow \Psi_I^-(o, p)]$ 
\item $\Psi_{P^{!+}} (o) = \inf_{p \in P} [\Psi_{P^+}(a) \rightarrow \Psi_I^-(o,a) ]$
\item $\Psi_{O^{?-}}(a) = \inf_{o \in O} [\Psi_{O^-}(o) \rightarrow \Psi_I^-(o, a)]$ 
\item $\Psi_{P^{!-}} (o) = \inf_{p \in P} [\Psi_{P^-}(a) \rightarrow \Psi_I^-(o,a) ]$
\end{itemize}

\noindent where $\rightarrow$ is fuzzy implication.  A fuzzy paraconsistent formal concept is then a pair where $O^{!+} = P^+, O^{!-} = P^-, P^{!+} = O^+, P^{!-} = O^-$.   Again, it appears that a concept system of this sort should form a complete lattice, but the proofs have not been worked out in full detail yet.

Defining concepts in this way yields concepts with fuzzy boundaries -- which is what is needed for naturalistic models of human-like concepts.  But these are paraconsistent fuzzy boundaries, which is precisely the situation considered in our treatment of the sorites paradox.   And when one parses through the mathematics, one finds that the paraconsistent foundations underlying the measurement of information content done to create pattern-sets directly connect to the paraconsistent foundations of the fuzziness of the concept boundaries.

There are also some interesting implications here for concept blending.   A blend of two concepts $C_1$ and $C_2$ is a new concept $C_3$ that takes some properties from $C_1$ and some from $C_2$ \cite{Fauconnier2002}.   The artistry of concept blending has to do with achieving interesting emergent properties from the juxtaposition of previously distinct properties.   What paraconsistent logic adds is the possibility that, when a property has truth value (1,0) in $C_1$ and (0,1) in $C_2$, it could be given any of the four CD truth values in $C_3$.   On the PLN level, this would result in $C_3$ having some property truth values $(C_3,p)$ selected from $(C_1, p)$ or $(C_2,p)$, and some determined via weighted-averaging the values $(C_1, p)$ and $(C_2,p)$.

\section{Future Directions}

While the investigation reported here is at the early stages, we have attempted to demonstrate the value of 4-valued p-bit based paraconsistent logic as a conceptual and formal foundation for a number of interrelated, critical AI ideas including PLN probabilistic logic and variations of probabilistic programming.  Our hope is that these ideas, as they are further developed, will be able to provide conceptual, mathematical and practical guidance regarding the development of cognitive algorithms, tools and systems, as well as perhaps aiding the understanding of human and other biological cognition.

\bibliographystyle{alpha}
\bibliography{bbm}

\end{document}